%% file: camera_ready.tex
\documentclass[letterpaper]{article} 
\usepackage{aaai2026}  
\usepackage{amsfonts} 
\usepackage{times}  
\usepackage{helvet}  
\usepackage{courier}  
\usepackage[hyphens]{url}  
\usepackage{graphicx} 
\usepackage{natbib}  
\usepackage{caption} 
\usepackage{algorithm}
\usepackage{algorithmic}

\usepackage{booktabs}
\usepackage{amsmath}
\usepackage{newfloat}
\usepackage{listings}
\usepackage{cleveref}
\usepackage{amssymb}

\newcommand{\xmark}{\ding{53}}
\DeclareCaptionStyle{ruled}{labelfont=normalfont,labelsep=colon,strut=off} 
\floatstyle{ruled}
\newfloat{listing}{tb}{lst}{}
\floatname{listing}{Listing}
\title{R-AVST: Empowering Video-LLMs with Fine-Grained Spatio-Temporal \\ Reasoning in Complex Audio-Visual Scenarios
}
\author{
    Lu Zhu\textsuperscript{\rm 1}\equalcontrib, 
    Tiantian Geng\textsuperscript{\rm 1, 2}\equalcontrib,
    Yangye Chen\textsuperscript{\rm 1},
    Teng Wang\textsuperscript{\rm 1, 3},
    Ping Lu\textsuperscript{\rm 4},
    Feng Zheng\textsuperscript{\rm 1, 5}\thanks{Corresponding author.}
}
\affiliations{
    \textsuperscript{\rm 1}Southern University of Science and Technology
    \textsuperscript{\rm 2}University of Birmingham\\
    \textsuperscript{\rm 3}The University of Hong Kong
    \textsuperscript{\rm 4}ZTE Corporation
    \textsuperscript{\rm 5}Spatialtemporal AI\\
    \{zlllllau, gengtiantian97\}@gmail.com, zhengf@sustech.edu.cn
}

\usepackage{bibentry}

\usepackage{multirow}
\usepackage{placeins}
\usepackage{booktabs}
\usepackage[table]{xcolor}
\usepackage{pifont}
\usepackage{bbding}
\usepackage{times}
\usepackage{helvet}
\usepackage{courier}
\usepackage{xcolor}

\usepackage[most]{tcolorbox}
\usepackage{amsmath}

\begin{document}
\maketitle

\begin{abstract}
\input{abstract}
\end{abstract}

\section{Introduction}
\input{introduction}

\section{Related Work}

\begin{figure*}[t]
\centering
\includegraphics[width=\textwidth]
{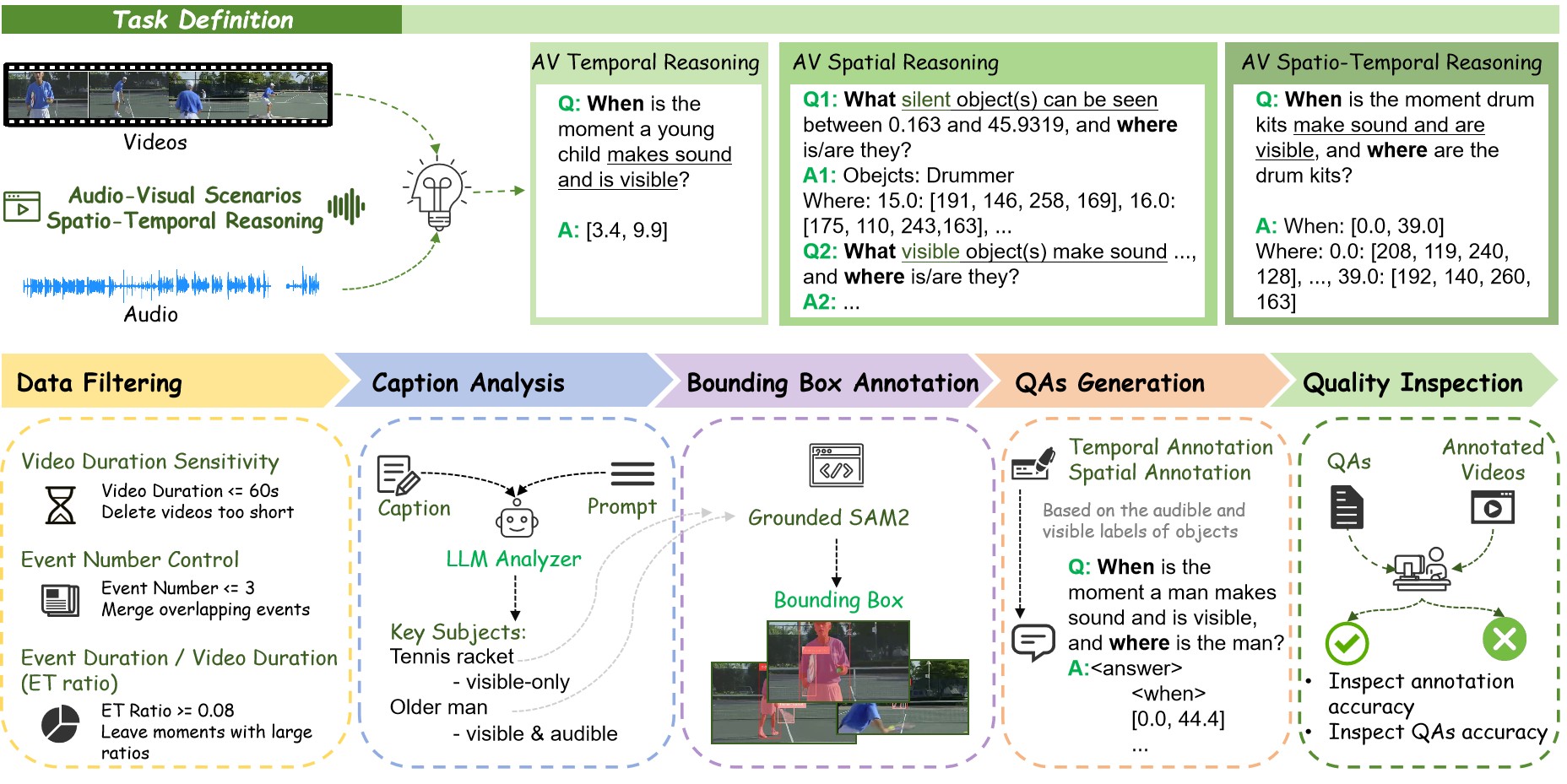}
\caption{Data generation pipeline of R-AVST. The tasks are explicitly designed to capture both spatial and temporal aspects of complex audio-visual scenes. The dataset construction follows a five-step process, yielding fine-grained spatio-temporal annotations and task-oriented QAs.}
\label{pipeline}
\end{figure*}
\input{relatedwork}

\section{R-AVST Dataset}
\input{dataset}
\section{AVST-Zero}
\input{method}

\section{Experiments}
\input{experiment}

\section{Conclusion}
We introduce R-AVST, a video dataset with fine-grained spatio-temporal annotations for complex audio-visual scenarios. Based on this dataset, we define three specialized reasoning tasks with automatically generated QAs. To support these tasks, we develop AVST-Zero, trained with GRPO and task-specific reward functions to improve spatio-temporal reasoning. Experimental results show that R-AVST advances research in audio-visual scenes, with our model demonstrating its effectiveness on these tasks.

\section{Acknowledgments}
This work was supported in part by the National Key Research and Development Program of China under Grant 2024YFE0203100, and in part by the ZTE Industry-University-Institute Cooperation Funds under Grant No.IA20240906004. 

\bibliography{ref}

\clearpage
\input{appendix}

\end{document}

%% file: abstract.tex
Recently, rapid advancements have been made in multimodal large language models (MLLMs), especially in video understanding tasks. However, current research focuses on simple video scenarios, failing to reflect the complex and diverse nature of real-world audio-visual events in videos.
To bridge this gap, we firstly introduce R-AVST, a dataset for audio-visual reasoning featuring fine-grained spatio-temporal annotations. In constructing this, we design a pipeline consisting of LLM-based key object extraction, automatic spatial annotation and manual quality inspection, resulting in over 5K untrimmed videos with 27K objects across 100 types of audio-visual events. Building on this dataset, we define three core tasks for spatio-temporal reasoning in audio-visual scenes and generate more than 8K high-quality, evenly distributed question-answer pairs to effectively benchmark model performance. To further enhance reasoning, we propose AVST-Zero, a reinforcement learning-based model that avoids intermediate supervision, directly optimizing behavior via carefully designed multi-dimensional rewards. Extensive experiments validate the effectiveness of our R-AVST in advancing audio-visual spatio-temporal reasoning, upon which AVST-Zero demonstrates competitive performance compared to existing models. To the best of our knowledge, R-AVST is the first dataset designed for real-world audio-visual spatio-temporal reasoning, and AVST-Zero offers a novel perspective for tackling future challenges in this domain.

%% file: introduction.tex
\begin{figure}[t]
\centering
\includegraphics[width=1\columnwidth]{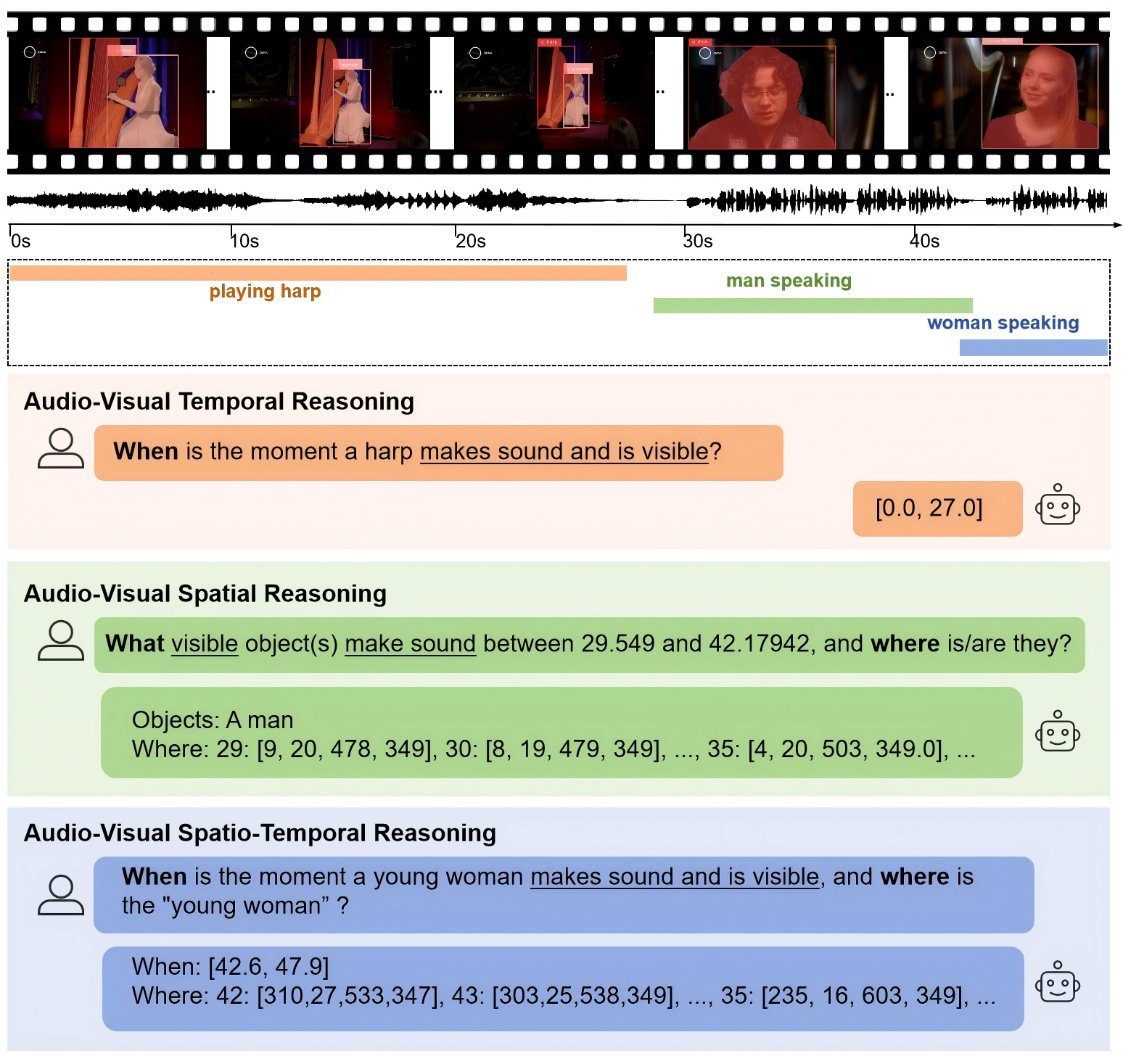}
\caption{Unlike previous datasets, R-AVST focuses on spatio-temporal reasoning in complex audio-visual scenes of untrimmed videos, offering fine-grained temporal boundary and spatial localization annotations. This example shows three core tasks designed to evaluate reasoning over sounding objects, time, and space.}
\label{introduction}
\end{figure}
The rapid advancement of multimodal large language models (MLLMs) recently has demonstrated their effectiveness in video understanding by integrating information from various modalities~\cite{wang2024comprehensive, zhang2024mm, tang2025video, geng2025longvale}. 
However, this rapid progress raises a question: \textit{Do current models and datasets adequately account for the compositional audio-visual nature of real-world video scenes, and leverage it to enhance spatio-temporal understanding of videos?} 
This issue is particularly important, as many videos originate from real-world audio-visual events, which play a vital role in practical applications, such as human-computer interaction, autonomous driving, and so on, where audio and visual modalities are both crucial for temporal and spatial perception.

In response to this issue, existing datasets such as AVE~\cite{tian2018audio}, UnAV-100~\cite{geng2023dense}, and PU-VALOR~\cite{tang2025empowering}, incorporate audio-visual information but primarily focus on temporal understanding while overlooking the spatial properties of audible objects. 
In contrast, spatio-temporal grounding datasets such as VidSTG~\cite{zhang2020does}, HC-STVG~\cite{tang2021human}, and V-STaR~\cite{cheng2025v} offer both temporal and spatial annotations, but they do not adequately capture the rich audio-visual dynamics of real-world scenes and typically involve only a limited range of object types. 
Therefore, proposing a fine-grained spatio-temporal dataset that jointly integrates audio and visual information is both valuable and urgent, as it can enhance models’ reasoning capabilities in real-world scenarios.

At the model level, models such as LLaVA-ST~\cite{li2025llava},  GroundingGPT~\cite{li2024groundinggpt} and Grounded-VideoLLM~\cite{wang2024grounded} have gradually extended their spatio-temporal modeling capabilities.
However, these models require extensive high-quality labeled data and lack sufficient exploration capacity.
With the promising performance of Deepseek-R1~\cite{guo2025deepseek} in rule-based reinforcement learning (RL), some studies have begun exploring the use of GRPO algorithms~\cite{shao2024deepseekmath} to enhance the reasoning capabilities of MLLMs. Models like VideoChat-R1~\cite{li2025videochat}, Video-R1~\cite{feng2025video} and Omni-R1~\cite{zhong2025omni} have emerged in this direction, yet they offer limited reward designs for spatio-temporal reasoning and lack dedicated tasks tailored for complex audio-visual scenarios.

To address the above limitations, we propose R-AVST, the first video dataset with fine-grained spatio-temporal annotations in complex audio-visual scenarios, as illustrated in Fig.~\ref{introduction}. 
In audio-visual scenarios, both auditory and visual attributes of objects can capture human perception and attention. Motivated by this, we use GPT-4o-mini~\cite{hurst2024gpt} to extract and analyze audio-visual event captions to label objects’ attributes. This allows models to determine whether an object is audible, visible, or both, and subsequently supports automatic spatial annotation for grounding object locations. To further assess models’ reasoning capabilities and align with human inference needs in complex audio-visual scenes, we define three targeted reasoning tasks, including temporal localization of sounding-visible objects, spatial localization of sounding-visible or silent-visible objects within a given duration, and spatio-temporal localization of sounding-visible objects. Based on these tasks, we automatically generate corresponding question-answer pairs (QAs) to enable large-scale evaluation. 
In total, R-AVST comprises 5,237 videos with 27,253 objects both sounding-visible and silent-visible and 8,166 QAs, covering over 100 types of audio-visual events, such as human speech, musical performances, and animal sounds.

Building upon R-AVST, we further focus on addressing the challenge of fine-grained spatio-temporal reasoning capabilities of models in complex audio-visual scenarios. 
To address the lack of advanced reasoning capabilities and the dependence on large-scale high-quality annotated data in models such as LLaVA-ST~\cite{li2025llava}, we fine-tune our model using the data-efficient GRPO~\cite{shao2024deepseekmath} method from DeepSeek-R1~\cite{guo2025deepseek}, given the rule-based characteristics of our tasks. Meanwhile, to overcome the absence of task-specific objectives and reward designs for complex audio-visual spatio-temporal reasoning in models like VideoChat-R1~\cite{li2025videochat}, we develop a multi-dimensional reward system tailored to our tasks. This system includes format, object, temporal, and spatial reward, which collectively enable effective policy gradient updates. Our experiments demonstrate that AVST-Zero achieves competitive performance on the three core tasks. It surpasses most Video-LLMs and sets a new perspective in the audio-visual spatio-temporal reasoning tasks.

Our contributions can be summarized as follows: 
\begin{itemize}
    \item We introduce R-AVST, the first video dataset encompassing a wide range of complex audio-visual events and featuring fine-grained spatio-temporal annotations, specifically designed to facilitate multimodal reasoning and evaluation in realistic scenarios of videos.
    \item Aiming to systematically evaluate models’ spatio-temporal reasoning capabilities and to align more closely with human retrieval demands in complex audio-visual contexts, we introduce three specialized tasks: Audio-Visual Temporal, Spatial, and Spatio-Temporal Reasoning, alongside automatically constructed QAs based on LLM-generated labels.
    \item We construct AVST-Zero, a Video-LLM fine-tuned in fully GRPO, trained on R-AVST to enhance its performance on audio-visual spatio-temporal reasoning tasks. Experimental results demonstrate that AVST-Zero achieves competitive performance across all three core tasks, validating its effectiveness.
\end{itemize}

%% file: relatedwork.tex
\subsection{Spatio-Temporal Understanding in Video-LLMs}
Spatio-temporal understanding is crucial for extracting key information from videos~\cite{goodge2025spatio, yuan2024video, zou2023beyond, geng2024uniav}. With the rise of MLLMs, general-purpose Video-LLMs like InternVL-2.5~\cite{chen2024expanding}, Qwen2.5-VL~\cite{bai2025qwen2}, VideoLLaMA3~\cite{zhang2025videollama}, and GroundingGPT~\cite{li2024groundinggpt} have made notable strides, along with specialized models such as LLaVA-ST~\cite{li2025llava}, VideoMolmo~\cite{ahmad2025videomolmo}, and Meerkat~\cite{chowdhury2024meerkat} that focus on spatio-temporal reasoning. These models benefit from improvements like stronger encoders and audio integration, yet still struggle with complex audio-visual spatio-temporal tasks.
On the dataset side, benchmarks like HC-STVG~\cite{tang2021human} and VidSTG~\cite{zhang2020does} provide spatio-temporal grounding annotations, but feature limited object diversity. While BOSTVG~\cite{yao2025omnistvg} introduces a multi-object setting but uses short videos. V-STaR~\cite{cheng2025v} targets long-video scenarios, yet lacks realistic audio-visual content. AV-UIE~\cite{du2025crab} centers on audio-visual understanding but mainly targets image-audio pairs, overlooking the construction of video-audio data.

\subsection{Reinforcement Learning Enhancement in LLMs}
Reinforcement Learning (RL) effectively enhances LLM reasoning with limited supervision. GPT-4~\cite{achiam2023gpt} uses PPO with reward and value functions, while ChatGLM3-DPO~\cite{glm2024chatglm} applies DPO through pairwise comparisons. DeepSeek-R1~\cite{guo2025deepseek} introduces GRPO~\cite{shao2024deepseekmath}, using rewards within groups to improve reasoning without value functions.
GRPO has since been adopted in Video-LLMs as a method after supervised fine-tuning (SFT). VideoChat-R1~\cite{li2025videochat} improves spatio-temporal perception, Video-R1~\cite{feng2025video} enhances temporal modeling, and R1-Omni~\cite{zhao2025r1} incorporates audio for emotion understanding.
Recent work~\cite{guo2025deepseek} explores fully RL-based training, as in Omni-R1~\cite{zhong2025omni}, which proposes an end-to-end GRPO framework. However, RL models specifically targeting spatio-temporal reasoning in audio-visual scenarios remain underexplored.

%% file: dataset.tex
\begin{figure*}[t]  
\centering
\includegraphics[width=\textwidth] 
{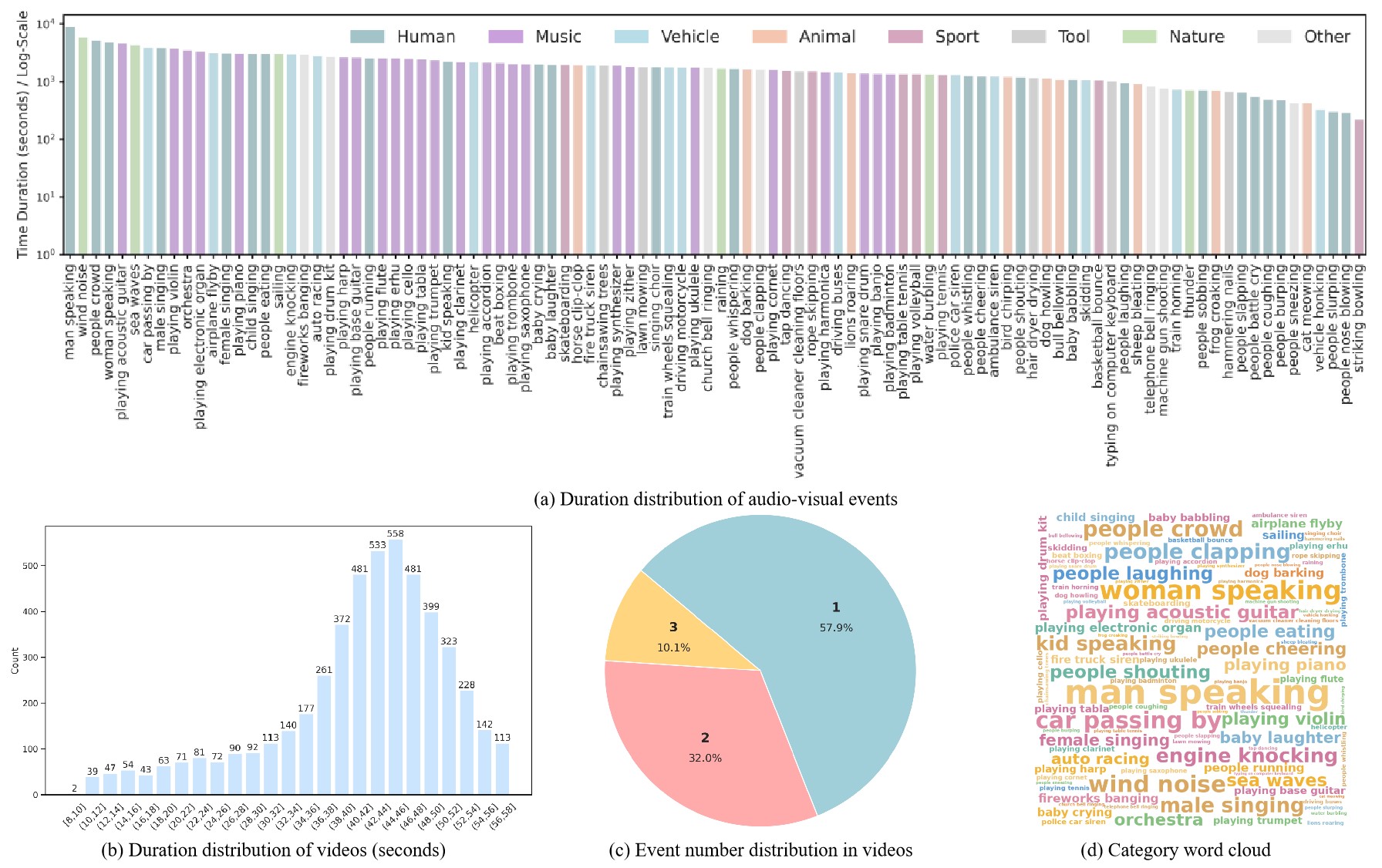}
\caption{Statistics of R-AVST dataset. (a) Duration distribution of different event categories in descending order, where colors represent their corresponding coarse-grained categories. (b) Duration distribution of all videos. (c) Distribution of the number of audio-visual events per video. (d) Word cloud of event categories.}
\label{overview}
\vspace{-2mm}
\end{figure*}

\begin{table}[t]
\centering
\setlength{\tabcolsep}{0.5mm} 
\small
\begin{tabular}{lccccccc}
\toprule
Dataset & \#Vid & \#Cls & Len & \#Obj & Mod & TB & SA \\
\midrule
Charades-STA & 10,009 & 157 & 30s     & -   & V  & \checkmark & \xmark \\
AVE     & 4,143  & 28   & 10s  & -   & VA  & \checkmark & \xmark \\
UnAV-100     & 10,790 & 100 & 42.1s   & -   & VA & \checkmark & \xmark \\
PU-VALOR     & 114,000  & -   & 10s  & -   & VA  & \checkmark & \xmark \\
LongVALE     & 8,411  & -   & 235s  & -   & VA  & \checkmark & \xmark \\
\midrule
VidSTG       & 6,924  & 79  & 28.01s  & 1   & V  & \checkmark & \checkmark \\
HCSTVG-v1    & 5,660  & 1   & 20s     & 1   & V  & \checkmark & \checkmark \\
HCSTVG-v2    & 16,544 & 1   & 20s     & 1   & V  & \checkmark & \checkmark \\
BOSTVG       & 10,018 & 23  & 36.5s   & 2.4 & V  & \checkmark & \checkmark \\
V-STaR       & 2,094  & -   & 110.23s & -   & V  & \checkmark & \checkmark \\
\midrule
AVSBench-V1  & 5,356  & 23  & 5s      & -   & VA & \xmark          & \checkmark \\
AVSBench-V2  & 12,356 & 70  & 7.64s   & -   & VA & \xmark          & \checkmark \\
VPO          & 22,019 & 21  & 10s     & -   & VA & \xmark          & \checkmark \\
LU-AVS       & 7,200  & 88  & 41.97s  & -   & VA & \xmark          & \checkmark \\
\midrule
\textbf{R-AVST (Ours)} & 5,237 & 100 & 42.17s & 5.2 & VA & \checkmark & \checkmark \\
\bottomrule
\end{tabular}
\caption{Comparison of R-AVST with previous related datasets. \#Cls: event category number; Len: average video duration; \#Obj: average video object number; Mod: modality of event captions; V: visual events; VA: audio-visual events; TB: temporal boundary; SA: spatial annotation.} 
\label{comparison}
\end{table}

\subsection{Task Definition}
Existing datasets mainly focus on visual-based temporal and spatial reasoning, often neglecting the audio modality. In complex real-world audio-visual contexts, auditory cues are crucial for accurately localizing objects in both time and space. However, the absence of fine-grained audio-visual scene annotations limits comprehensive evaluation, as shown in Tab.~\ref{comparison}. 
To fill this gap, we propose R-AVST dataset and three reasoning tasks designed for spatio-temporal reasoning in complex audio-visual scenarios, as Fig.~\ref{pipeline} shows. 

\paragraph{Audio-Visual Temporal Reasoning}
How to ground the temporal period when an  object makes sound and is visible helps us better extract key information from audio-visual video scenarios. In this dimension, we design the Audio-Visual Temporal Reasoning task to infer about the time when the object appears and makes sound.

\paragraph{Audio-Visual Spatial Reasoning}
To better capture the spatial relationships among objects in audio-visual events, we introduce a novel Audio-Visual Spatial Reasoning task that considers both sound-emitting and non-sound-emitting objects. Given a specific temporal segment of an audio-visual event, Video-LLMs are supposed to accurately ground the target objects within the scene.

\paragraph{Audio-Visual Spatio-Temporal Reasoning}
In real-world scenarios, human perception of the audio-visual events is usually a process of simultaneously obtaining temporal and spatial information. Therefore, Audio-Visual Spatio-Temporal Reasoning task aims to evaluate the joint understanding ability of the model for temporal and spatial information in audio-visual scenes, making it closer to the real perception mechanism of human beings. Specifically, given that an object is known to have both visible and audible state, the task objective is to identify the temporal period when the object appears and further locate its spatial position within these periods.

\subsection{Dataset Construction}
\paragraph{Data Collection and Filtering}

We collect videos from UnAV-100~\cite{geng2023dense}, an audio-visual dataset of untrimmed videos covering diverse domains. Starting from raw YouTube videos and the corresponding event captions, we employ a filtering strategy to select high-quality samples, aiming to ensure comprehensive coverage of complex audio-visual scenarios. As detailed in Fig.~\ref{pipeline}, the filtering involves three steps to balance video duration, event count, and event coverage ratio. First, videos are grouped into short (0-20s), medium (20-40s), and long (40-60s) durations, excluding extremely short clips and merging overlapping events. Second, we limit videos to at most three audio-visual events, ensuring a balanced distribution of 1, 2, and 3-event videos to facilitate clearer scene-level evaluation. Third, videos with an event-to-total (ET) duration ratio below 0.08 are removed, retaining those where meaningful events occupy a substantial portion.
Finally, we curate a dataset of 5,237 high-quality videos, encompassing a wide variety of complex audio-visual scenarios.

\paragraph{Caption Analysis}

Audio-visual spatio-temporal reasoning tasks typically focus on objects, mostly expressed as nouns in captions. We use GPT-4o-mini~\cite{hurst2024gpt} as an Analyzer LLM to extract noun-based objects from captions, as shown in Fig.~\ref{pipeline}. To capture the multi-modal nature, tailored prompts guide the model to annotate each object's auditory and visual attributes in a standardized format. We query the analyzer by emphasizing the definition of audibility in object-sound relations to improve analysis accuracy. For example, in the caption ``A group of people are sailing on silver sailboats'', ``a group of people'' is labeled ``visible\&audible'' while ``silver sailboats'' is ``visible-only''.
Finally, we identify 27,253 objects in audio-visual event captions in total, with 50.88\% labeled in ``visible\&audible''.

\paragraph{Bounding Box Annotation}

Based on caption analysis and temporal annotations, we perform spatial annotation on video frames within audio-visual event segments. To reduce the high annotation cost of large-scale videos, we leverage the automatic tool, Grounded-SAM2~\cite{ravi2024sam}, for fine-grained, frame-by-frame object annotation.
The pipeline extracts frames corresponding to each event and constructs textual prompts based on object information derived from caption analysis, which are then fed into Grounded-SAM2~\cite{ravi2024sam}.

\paragraph{Automatic QAs Generation}
To align with the three reasoning tasks we propose for audio-visual scenes, R-AVST defines three corresponding question types: \textbf{when} (temporal), \textbf{where} (spatial), and \textbf{what} (object). Below are question detailed settings for each task:
\begin{itemize}
    \item Audio-Visual Temporal Reasoning: 
With certain visible and audible objects, the question is formulated as: \textit{When is the moment [objects] make sound and are visible?}
    \item Audio-Visual Spatial Reasoning:
Given a time interval, the questions are formulated as: (1) For sounding-visible objects: \textit{What objects make sound between [start\_time] and [end\_time], and where are they?} (2) For silent-visible objects: \textit{What silent objects can be seen between [start\_time] and [end\_time], and where are they?}
    \item Audio-Visual Spatio-Temporal Reasoning:
With certain visible and audible objects, the question is formulated as: \textit{When is the moment [objects] make sound and are visible, and where are they?}
\end{itemize}
The answer follows a unified format, with different tags depending on the question type. Programs extract object audio-visual labels and generate corresponding QAs. The training set contains 2,663 temporal, 2,666 spatial, and 1,204 spatio-temporal questions, while the test set includes 663, 664, and 306 questions of each type, respectively.
\paragraph{Quality Control}

To ensure the high quality of the R-AVST dataset, we conduct manual verification of the spatio-temporal annotations and QAs at this stage. Videos with incorrect spatio-temporal annotations or inaccurate QAs are removed. This process enhances the robustness and accuracy of the R-AVST dataset, providing a more reliable basis for further evaluating of Video-LLMs.

\subsection{Dataset Analysis}

\paragraph{Overview}
Overall, we introduce R-AVST, the first dataset specifically designed to evaluate the spatio-temporal reasoning capabilities of Video-LLMs in complex audio-visual scenarios. As illustrated in Fig.~\ref{overview}, the dataset comprises 5,237 videos of 100 categories for over 220,833 seconds, with an average duration of 42.17 seconds per video. 
The training/test split has 4,171/1,066 videos with 6,533/1,633 QAs, respectively.
These videos encompass a wide range of audio-visual events, with a relatively balanced distribution across videos containing 1, 2, or 3 events. Importantly, each video is accompanied by fine-grained spatio-temporal annotations, providing strong support for advancing research in reasoning within complex audio-visual video contexts.

\paragraph{Comparisons with Existing Datasets}

As shown in Tab.~\ref{comparison}, we compare R-AVST with existing relevant datasets~\cite{gao2017tall,tian2018audio, tang2025empowering, geng2023dense, geng2025longvale, zhang2020does,tang2021human,yao2025omnistvg,cheng2025v,zhou2022audio,zhou2025audio,liu2024benchmarking,chen2024unraveling}. Existing datasets primarily focus on general temporal or visual grounding scenarios and lack fine-grained modeling of audio-visual scenarios. On the other hand, while datasets such as AVSBench~\cite{zhou2022audio,zhou2025audio}, VPO~\cite{chen2023closer}, and LU-AVS~\cite{liu2024benchmarking} focus on spatial localization of audible objects, they often neglect the temporal aspect. In contrast, R-AVST covers a wider range of events and objects, and is tailored for fine-grained spatio-temporal reasoning in real-world audio-visual scenes, offering a more comprehensive benchmark for complex video reasoning tasks.

%% file: method.tex
To further enhance the spatio-temporal reasoning capabilities of Video-LLMs, we fine-tune the model with GRPO~\cite{shao2024deepseekmath}, along with task-oriented rewards designed to improve performance on the audio-visual spatio-temporal reasoning domain.
\begin{figure}[t]
\centering
\includegraphics[width=\columnwidth]{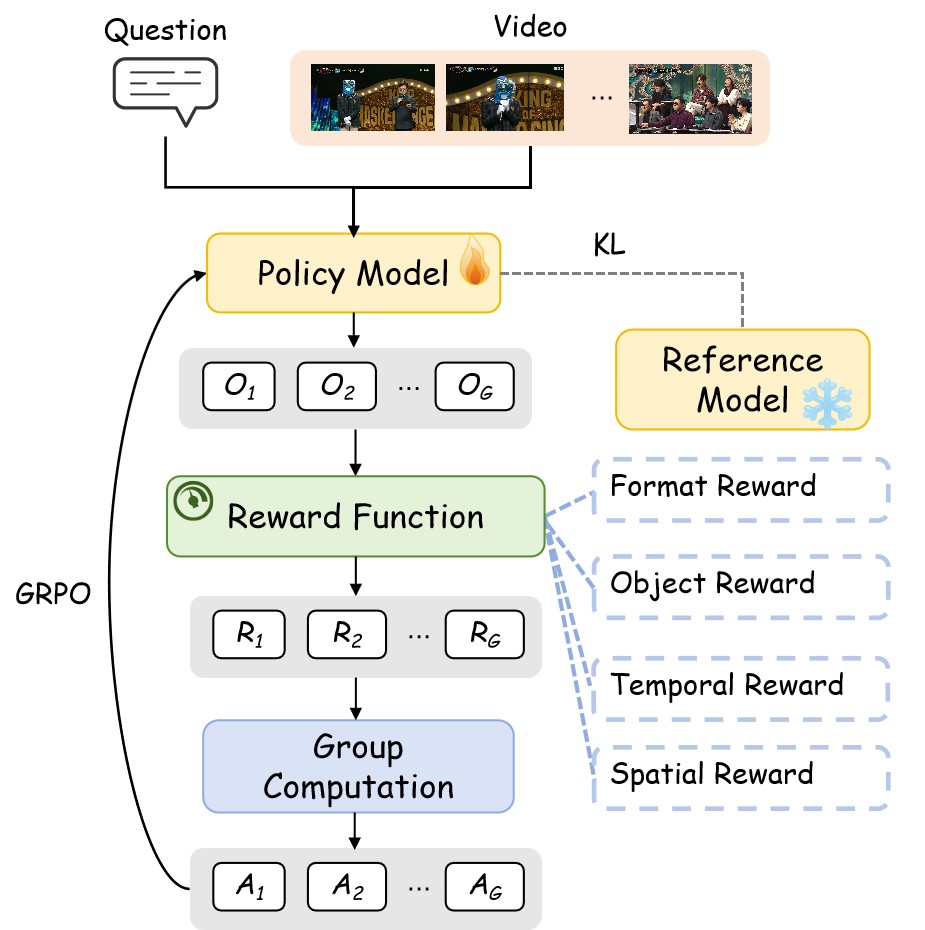} 
\caption{Model architecture of our AVST-Zero model. The multi-dimensional reward design allows AVST-Zero to perform exceptionally well in spatio-temporal reasoning tasks. }
\label{model}
\end{figure}
\subsection{Group Relative Policy Optimization}
GRPO is a reinforcement learning (RL) method and can be used in multimodal large language models (MLLMs) to guide task alignment through reward functions. It differs from PPO in that it reduces reliance on a critic model by directly comparing groups of generated responses. Specifically, for each question $q$, GRPO~\cite{shao2024deepseekmath} samples a set of outputs $\left\{o_1, o_2, \ldots, o_e\right\}$ from the old policy $\pi_{\theta_{\text {old }}}$, and then optimizes the policy $\pi_\theta$ by maximizing the objective defined as:
\begin{align}
\mathcal{J}_{\text{GRPO}}(\theta) ={}& \mathbb{E}\bigg[
q \sim P(Q),\, \{o_i\}_{i=1}^G \sim \pi_{\theta_{\text{old}}}(O \mid q)
\bigg] \notag \\
& \frac{1}{G} \sum_{i=1}^G \bigg(
\min\bigg(
\frac{\pi_\theta(o_i \mid q)}{\pi_{\theta_{\text{old}}}(o_i \mid q)} A_i, \notag \\
&\quad \operatorname{clip}\left(
\frac{\pi_\theta(o_i \mid q)}{\pi_{\theta_{\text{old}}}(o_i \mid q)},
1 - \varepsilon,
1 + \varepsilon
\right) A_i
\bigg) \notag \\
&\quad - \beta \, \mathbb{D}_{\text{KL}}(\pi_\theta \| \pi_{\text{ref}})
\bigg),
\label{GRPO}
\end{align}
where $A_i$ denotes the relative advantage of the $i$-th sample within the group of generated responses, estimated directly from rule-based rewards $\left\{r_1, r_2, \ldots, r_G\right\}$. The term $\mathbb{D}_{K L}\left(\pi_\theta \| \pi_{r e f}\right)$ denotes the KL divergence, which measures the degree to which the optimized policy model $\pi_\theta$ deviates from the reference model $\pi_{\mathrm{ref}}$. 
The hyper-parameters $\varepsilon$ and $\beta$ control the clipping threshold of the advantages and the penalty intensity of the KL-regularization term, respectively.

\subsection{Rewards Design}
To align with the specific characteristics of our tasks, we introduce four distinct reward components: format, object, temporal, and spatial. Each is designed to capture and reinforce a particular dimension of the tasks.

\paragraph{Format Reward}
The reward $R_{\text{format}}$ assesses the format consistency of outputs by checking whether the required tag pairs (\texttt{<answer>}, \texttt{<object>}, \texttt{<when>}, \texttt{<where>}) are correctly included and matched, based on task requirements.

\paragraph{Object Reward}
We use Word2Vec~\cite{mikolov2013efficient} to quantify the semantic similarity between the predicted and ground truth object names, which helps mitigate misclassification caused by lexical variations. The similarity $\operatorname{sim}\left(V_{\text{pred}}, V_{\text{gt}}\right)$ is defined as:
\begin{equation}
\operatorname{sim}\left(V_{\text{pred}}, V_{\text{gt}}\right)
= \frac{V_{\text{pred}} \cdot V_{\text{gt}}}{\left\|V_{\text{pred}}\right\| \left\|V_{\text{gt}}\right\|},
\label{sim}
\end{equation}
where $V_{\text{pred}}$ and $V_{\text{gt}}$ denote the word embeddings of the predicted and ground truth object names, respectively. 
Based on this, the object reward is defined as follow, where $\tau$ is the similarity threshold.
\begin{equation}
R_{\text{object}}= \begin{cases}1, & \text{if} \operatorname{sim}\left(V_{\text{pred}}, V_{\text{gt}}\right) \geq \tau. \\ 0, & \text{otherwise.}\end{cases}
\label{obj_reward}
\end{equation}

\paragraph{Temporal Reward}
To enhance the model’s temporal reasoning ability, this reward evaluates the accuracy of predicted audio-visual event segments by measuring the overlap between the predicted interval $I_{\text{pred}}$ and the ground truth interval $I_{\text{gt}}$. The temporal reward computation is defined as their intersection-over-union (IoU) ratio: 
\begin{equation}
R_{\text {temporal }}=\frac{\left|I_{\text {pred}} \cap I_{\text {gt}}\right|}{\left|I_{\text {pred}}\cup I_{\text{gt}}\right|}.
\label{temporal_reward}
\end{equation}

\paragraph{Spatial Reward}

To enhance fine-grained spatial reasoning, we define a spatial reward as the average 2D IoU between predicted and ground truth bounding boxes over their overlapping temporal interval. For each time point \( t \in [T_{\text{start}}, T_{\text{end}}] \), the IoU is computed as follow if the object prediction is correct:
\begin{equation}
\text{IoU}(t) = \frac{\text{Area}(B_{\text{pred}}(t) \cap B_{\text{gt}}(t))}{\text{Area}(B_{\text{pred}}(t) \cup B_{\text{gt}}(t))}.
\label{sr_t}
\end{equation}
Therefore, the spatial reward \( R_{\text{spatial}} \) can be computed as the mean IoU over all \( N \) time points:
\begin{equation}
R_{\text{spatial}} = \frac{1}{N} \sum_{t=T_{\text{start}}}^{T_{\text{end}}}
\text{IoU}(t).
\label{sr}
\end{equation}

\paragraph{Final Reward}
The total reward is computed by the weighted sum of the rewards in different parts as:  
\begin{equation}
R=\lambda_{\text {f}} R_{\text {format}}+\lambda_{\text{t}} R_{\text{temporal}}+\lambda_{\text {o}} R_{\text {object}}+\lambda_{\text{s}}R_{\text {spatial}}.
\label{R}
\end{equation}
For all tasks, $\lambda_{\text{f}} = 1$, while other parameters ($\lambda_{\text{t}}$, $\lambda_{\text{o}}$, $\lambda_{\text{s}}$) are set depending on the task type.

%% file: experiment.tex
\subsection{Experimental Settings}
\begin{table*}[t]
\centering
\small
\setlength{\tabcolsep}{0.7mm}
\begin{tabular}{l|c|ccc|cccc|ccc}
\toprule
\multirow{3}{*}{\textbf{Method}} & \multicolumn{4}{c|}{\textbf{Audio-Visual Spatial Reasoning}} & \multicolumn{7}{c}{\textbf{Audio-Visual Spatio-Temporal Reasoning}} \\
& \multicolumn{1}{c}{Object} & \multicolumn{3}{c|}{Spatial} & \multicolumn{4}{c}{Temporal} & \multicolumn{3}{c}{Spatial} \\
\cmidrule(lr){2-5} \cmidrule(lr){6-12}
& Accuracy & m\_vIoU & AP@0.3 & AP@0.5 & m\_tIoU & R1@0.3 & R1@0.5 & R1@0.7 & m\_vIoU & AP@0.3 & AP@0.5 \\
\midrule
Qwen2.5-VL(7B)   & 1.91 & \underline{2.31} & 0.90 & 0.15 & 34.55 & 45.10 & 29.74 & 15.69 & 1.37 & 1.14 & 0.65 \\
Qwen2.5-Omni(7B) & 14.04 & 1.96 & 2.43 & \underline{1.24} & 33.44 & 43.46 & 19.28 & 9.80 & 2.85 & 2.99 & 1.04 \\
Video-LLaMA3(7B) & 15.07 & 1.27 & 0.16 & 0.15 & 37.43 & 50.65 & 34.64 & 22.22 & 1.69 & 1.63 & 0.00 \\
GroundingGPT(7B) & 0.55 & 0.16 & 0.00 & 0.00 & 13.65 & 15.05 & 6.02 & 1.67 & 5.59 & 3.68 & 0.00 \\
InternVL2.5(8B)  & \underline{15.97} & 0.74 & 0.66 & 0.00 & 21.46 & 26.47 & 13.07 & 7.52 & 2.87 & 2.80 & 0.00 \\
Video-R1(7B)     & 13.02 & 1.19 & 0.95 & 0.40& 22.05 & 26.47 & 10.13 & 5.23 & 0.15 & 0.11 & 0.00 \\
VieoChat-R1(7B)  & 15.54 & 1.99 & 3.11 & 0.36 & \underline{41.81} & \underline{60.78} & \underline{44.77} & \textbf{25.49} & 2.15 & 3.21 & 0.60 \\
\midrule
\textbf{AVST-Zero (7B)} & 14.34 & 2.27 & \underline{3.12} & 0.87 & \textbf{46.04} & \textbf{67.32} & \textbf{46.08} & \underline{23.53} & \underline{8.59} & \underline{10.38} & \underline{3.83} \\
\textbf{AVST-Zero-Omni (7B)} & \textbf{19.48} & \textbf{3.87} & \textbf{4.47} & \textbf{2.17} & 35.97 & 50.00 & 20.92 & 10.46 & \textbf{17.74} & \textbf{22.90} & \textbf{12.26} \\
\bottomrule
\end{tabular}
\caption{Comparison of different Video-LLMs for audio-visual spatial and spatio-temporal reasoning tasks on R-AVST test set.}
\label{exp_res1}
\end{table*}

In experiments, we base Qwen2.5-VL 7B~\cite{bai2025qwen2} and Qwen2.5-Omni 7B~\cite{xu2025qwen2} to fine-tune two model variants on our R-AVST dataset, namely AVST-Zero and AVST-Zero-Omni, respectively.  
Training is conducted on four NVIDIA RTX A6000 GPUs for a single epoch, with a batch size of 1 on each device. The group generation number is set to 6. 

\subsection{Quantitative Results}

\begin{table}[t]
\centering
\small
\setlength{\tabcolsep}{0.8mm}
\begin{tabular}{l|cccc}
\toprule
\multirow{2}{*}{\textbf{Method}} & \multicolumn{4}{c}{\textbf{Audio-Visual Temporal Reasoning}} \\
\cmidrule(lr){2-5}
                       & m\_tIoU & R1@0.3 & R1@0.5 & R1@0.7 \\
\midrule
Qwen2.5-VL(7B)         & 36.05 & 46.40 & 34.38 & 16.22 \\
Qwen2.5-Omni(7B)       & 30.70 & 37.09 & 18.92 & 9.01  \\
Video-LLaMA3(7B)       & 37.17 & 50.30 & 35.29 & 22.67 \\
GroundingGPT(7B)       & 10.77 & 11.06 & 5.38  & 1.84  \\
InternVL2.5(8B)        & 18.37 & 22.07 & 10.21 & 5.26  \\
Video-R1(7B)           & 22.48 & 22.82 & 12.16 & 6.01  \\
VieoChat-R1(7B)        & \underline{43.17} & \underline{60.81} & \underline{46.70} & \textbf{25.68} \\
\midrule
\textbf{AVST-Zero (7B)} & \textbf{47.96} & \textbf{71.13} & \textbf{51.43} & \underline{23.91} \\
\textbf{AVST-Zero-Omni (7B)} & 34.79 & 51.05 & 21.32 & 11.56 \\
\bottomrule
\end{tabular}
\caption{Comparison of different Video-LLMs for audio-visual temporal reasoning tasks on R-AVST test set.}
\label{exp_res2}
\end{table}

As shown in Tab.~\ref{exp_res1} and Tab.~\ref{exp_res2}, we compare our models with existing advanced models~\cite{bai2025qwen2, xu2025qwen2, cheng2024videollama,li2024groundinggpt,chen2024expanding,li2025videochat,feng2025video}.
The evaluation is conducted on the R-AVST test set. 
In Tab.~\ref{exp_res1}, for the audio-visual spatial reasoning task, AVST-Zero performs similarly to other models in m\_vIoU but slightly outperforms them in AP@0.3 with 3.12\%. For the audio-visual spatio-temporal reasoning task, AVST-Zero shows significantly better performance, achieving 46.04\% m\_tIoU and 8.59\% m\_vIoU. As shown in Tab.~\ref{exp_res2}, AVST-Zero leads in average temporal perception for the audio-visual temporal reasoning task with an m\_tIoU of 47.96\%.
We also observe that AVST-Zero-Omni achieves higher prediction accuracy than AVST-Zero in the object and spatial dimensions, while performing worse in the temporal dimension. This is attributed to the base model’s strong audio–visual joint perception but relatively weak temporal perception capabilities.
These results show that our AVST-Zero variants perform competitively across all three tasks, confirming their effectiveness.

\subsection{Ablation Study}
\begin{table}[h]
\centering
\small
\setlength{\tabcolsep}{0.7mm}
\begin{tabular}{l|c|cc|cc}
\toprule
\multirow{2}{*}{\textbf{Model}} & \textbf{AVTR} & \multicolumn{2}{c|}{\textbf{AVSR}} & \multicolumn{2}{c}{\textbf{AVSTR}} \\
                       & m\_tIoU & Obj Acc & m\_vIoU & m\_tIoU & m\_vIoU \\
\midrule
SFT                    & 42.84   & 9.52   & 3.42   & 38.40   & 4.26   \\
AVST-Zero           & \textbf{48.17}   & 20.72   & \textbf{4.62}    & \textbf{46.93}   & \textbf{10.87}   \\
\begin{tabular}[c]{@{}l@{}}AVST-Zero\\ (w/o temporal reward)\end{tabular}
                       &     46.67    &  \textbf{23.95}       &   \underline{4.54}     &  \underline{45.82}       &    8.31     \\
\begin{tabular}[c]{@{}l@{}}AVST-Zero\\ (w/o spatial reward)\end{tabular}
                       & \underline{47.03}  & \underline{23.17}   & 3.28    & 44.29   & \underline{9.23}    \\
\bottomrule
\end{tabular}
\caption{Ablation study on supervised fine-tuning (SFT) and the reward components. AVTR, AVSR, and AVSTR denote the Audio-Visual Temporal, Spatial, and Spatio-Temporal Reasoning tasks, respectively.}
\label{ablation}
\vspace{-4mm}
\end{table}

We conduct ablation experiments on a test subset with a balanced ratio 1:1:1 of the three task types to comprehensively assess the impact of each reward. As shown in Tab.~\ref{ablation}, removing the temporal reward reduces temporal accuracy from 48.17\% to 46.67\%, while removing the spatial reward significantly lowers m\_vIoU, from 4.62\% to 3.28\% (spatial reasoning) and from 10.87\% to 9.23\% (spatio-temporal reasoning). Meanwhile, we also observe that the interdependence of the spatio-temporal dimension leads to cross-effects between different reward modules. 
Moreover, compared with simple SFT, directly applying RL yields more substantial benefits across all three tasks, suggesting that RL is better suited to the tasks' fine-grained nature.

\subsection{Qualitative Results}
\begin{figure}[t]
\centering
\includegraphics[width=.95\columnwidth]{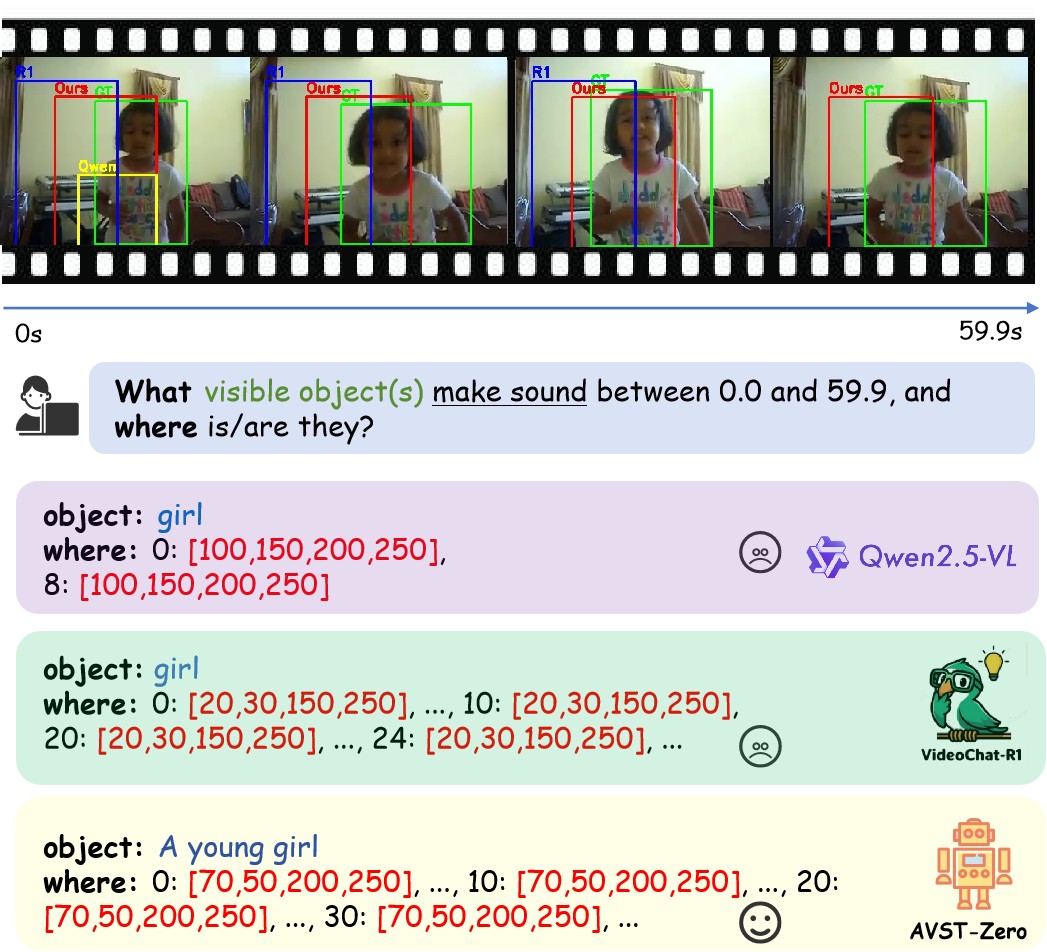}
\caption{Qualitative results. For the bounding boxes in the video: green denotes the ground truth, blue comes from VideoChat-R1, yellow from Qwen2.5-VL, and red from our AVST-Zero.}
\label{result}
\vspace{-2mm}
\end{figure}

As shown in Fig.~\ref{result}, Qwen2.5-VL~\cite{bai2025qwen2} predicts sparse and inaccurate object locations. VideoChat-R1~\cite{li2025videochat} correctly identifies the girl, but our model yields results closer to the ground truth, with more accurate object recognition and spatial localization. 

%% file: appendix.tex
\section{Appendix}

\subsection{More Implementation Details}
\paragraph{Bounding Box Annotation}
Due to the high cost and inefficiency of manually annotating large-scale video data, we introduce an automated annotation pipeline based on Grounded-SAM2~\cite{ravi2024sam} to perform fine-grained, frame-by-frame object spatio-temporal annotation in complex audio-visual scenarios. In this automated process, we first extract video frames corresponding to the audio-visual events, and then construct text prompts using object names identified during the caption analysis stage. These prompts are used as inputs to Grounded-SAM2~\cite{ravi2024sam}.
For the configuration, we set the BOX\_THRESHOLD to 0.4 and the TEXT\_THRESHOLD to 0.3, representing the minimum confidence score for bounding boxes and the minimum similarity score between output text labels and input prompts, respectively.
Finally, we obtain bounding box annotations for each frame, which are stored in a structured JSON format.

\paragraph{Human Evaluation}
We employ a dual-human scoring procedure using a 4-point scale for the test dataset to ensure data quality. The inter-rater reliability, measured by Quadratic Weighted Cohen’s $k$, reaches 0.71 on a subset, indicating substantial agreement. Based on this process, we filter 1,066 videos from the original 2,131 videos for the final test set. Subsequently, quantitative evaluation is conducted using Video-LLMs with the same 1-4 scoring scale. The results show that the mean score for the original test dataset is 2.54, while the curated test dataset achieves a higher mean score of 2.93, suggesting improved overall quality and consistency.

The scoring criteria are defined as follows:
\begin{itemize}
    \item 4: Excellent-Fully accurate and consistent with both visual and audio content.
    \item 3: Good-Mostly accurate, with minor mismatches or omissions.
    \item 2: Fair-Partially correct, with several inconsistencies or missing details.
    \item 1: Poor-Largely inaccurate or inconsistent with the video.
\end{itemize}

\paragraph{AVST-Zero Training}
We set the \textit{max\_prompt\_length} to 512 tokens and the \textit{max\_completion\_length} to 1024 tokens. To constrain the visual resolution, we set \textit{max\_pixels} to 12,845,056 and \textit{min\_pixels} to 3,136. 

\paragraph{Evaluation Metrics}
Following prior works~\cite{yao2025omnistvg,wang2025spacevllm,cheng2025v}, we adopt evaluation metrics across different dimensions. At the object level, prediction accuracy is determined by the proportion of cases where the similarity score surpasses a given threshold, as specified in Eq.~\ref{sim} and Eq.~\ref{obj_reward}. At the temporal level, we evaluate performance using the mean temporal Intersection over Union (m\_tIoU) and Recall@1 at tIoU thresholds of \{0.3, 0.5, 0.7\}. At the spatial level, we evaluate performance using the mean visual IoU (m\_vIoU) by frames, and the Average Precision Score (AP)@1 at vIoU thresholds of \{0.3, 0.5\}.

\subsection{Dataset Summary}
\begin{figure}[h]
\centering
\includegraphics[width=0.5\columnwidth]{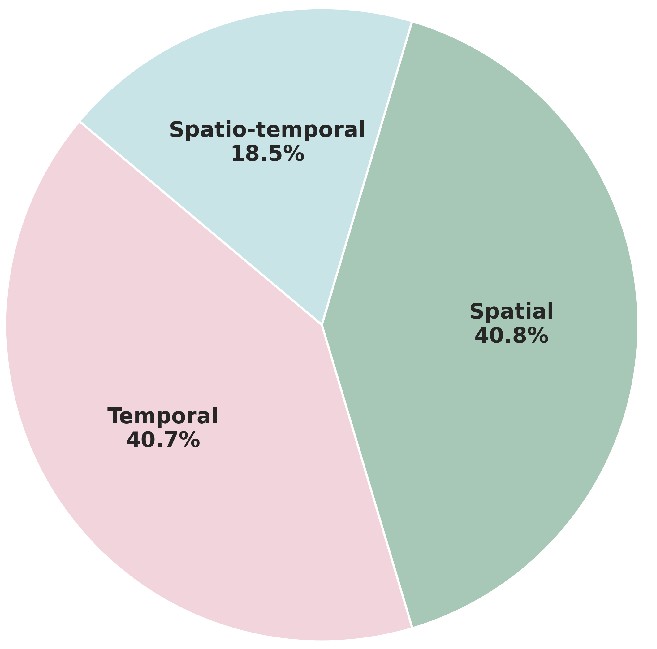}
\caption{Distribution of Temporal, Spatial, and Spatio-Temporal QAs in R-AVST. }
\label{QAdistribution}
\end{figure}

This section presents the distribution of the training and test sets in the R-AVST dataset as Tab.~\ref{split} shows, including the distribution of temporal, spatial, and spatio-temporal question-answer pairs.
To better characterize complex audio-visual scenes, we aim for a relatively balanced distribution across different task types as Fig.~\ref{QAdistribution} shows.
\begin{table}[h]
\centering
\begin{tabular}{lccc}
\toprule
\textbf{Split} & \textbf{Temporal} & \textbf{Spatial} & \textbf{Spatio-Temporal} \\
\midrule
\textit{Train} & 2663 & 2666 & 1204 \\
\textit{Test}  & 663  & 664  & 306  \\
\midrule
\textbf{Total} & 3326 & 3330 & 1510 \\
\bottomrule
\end{tabular}
\caption{Number of QAs in each split across train and test dataset.}
\label{split}
\end{table}

\subsection{Prompt Template}
\paragraph{Prompt Template for Caption Analysis}
We employ GPT-4o-mini~\cite{achiam2023gpt} as our caption analyzer to extract objects and their audio-visual attributes from the audio-visual event captions. As shown in the Box 1, we refine the definition of audible elements and enhance the association between subject and action to guide the model toward more accurate understanding.

\begin{tcolorbox}[breakable,
                  colback=white!100!white,
                  colframe=gray!150!black,
                  title=Box 1: Prompt Template for Caption Analysis.,
                  fonttitle=\bfseries,
                  coltitle=black,
                  boxrule=0.8pt,
                  arc=2mm]
You are an AI assistant specialized in caption component analysis.
Given a caption, complete the following tasks:

\textbf{1. Key Subject Identification}

Identify the key subjects mentioned in the caption.

Definition:\\
Key subjects are dominant, dynamic, or sound-producing (and visible) objects that play a leading role in the event described by the caption.

Important Principles:\\
A key subject must be a concrete entity (agent) that can be meaningfully associated with an action or sound. For example, in ``A person is shouting," the key subject is ``a person", not ``shouting".

Verbs or actions are not subjects unless they can be directly attributed to a specific agent.

Prioritize:\\
Dynamic entities (e.g., people, animals, moving vehicles),\\
Visible \& sound-producing entities,\\
Include static or silent entities only if no dynamic/sound-producing entities exist.

\textbf{2. Labeling}

Assign one label to each key subject:\\
visible-only: If the entity is only visible and not producing sound.\\
visible \& audible: If the entity is both visible and producing sound.\\
Do not assign audible-only to vague or environmental sounds (like ``noise", ``shouting").

\textbf{3. Output Format}

Present the analysis in the following structured format:

Key Subjects:

[Entity Name] - [Label: visible-only / audible-only / visible \& audible]

[Entity Name] - [Label: visible-only / audible-only / visible \& audible]

...

Subject Number: [Number of key subjects]

Important Instructions:

Base your analysis strictly on the caption content.
Do not assume or infer based on background knowledge or unstated context.
Avoid speculative interpretation. Stick to explicitly mentioned or clearly implied entities.
Ensure formatting is clean, concise, and consistent.

Keep the output clean, concise, and consistently formatted.
\label{analyzer}
\end{tcolorbox}

\paragraph{Prompt Template for AVST-Zero Training and Inference}
During GRPO-based training and model inference, we design a prompt as shown in Box 2, to guide the model. 
The \texttt{\textless answer\textgreater} tag is used to obtain the full response, 
while \texttt{\textless what\textgreater}, \texttt{\textless when\textgreater}, and \texttt{\textless where\textgreater} 
are employed to extract the predicted object, temporal span, and spatial information at each timestamp, respectively.

\begin{tcolorbox}[breakable,
                  colback=white!100!white,
                  colframe=gray!150!black,
                  title=Box 2: Prompt Template for AVST-Zero Training and Inference.,
                  fonttitle=\bfseries,
                  coltitle=black,
                  boxrule=0.8pt,
                  arc=2mm]
You are an expert in audio-visual video grounding.

For each question, output within \textless answer\textgreater\textless /answer\textgreater:\\
- \textless object\textgreater: grounded object(s), e.g. `man', `dog', separated by commas\\
- \textless when\textgreater: [start, end] in seconds, 1 decimal\\
- \textless where\textgreater for each object by each second within the object appearing duration:\\
\hspace*{4mm}timestamp: [x1, y1, x2, y2]

Only include tags \textless object\textgreater, \textless when\textgreater and \textless where\textgreater. No extra text.

Example: \\
\textless answer\textgreater\\
\textless when\textgreater[10.0,20.5]\textless /when\textgreater\\
\textless object\textgreater dog \textless /object\textgreater\\
\textless where\textgreater\\
10.0: [100,200,300,400]\\
11.0: [109,280,320,432]\\
12.0: [100,200,300,400]\\
\textless /where\textgreater\\
\textless object\textgreater cat \textless /object\textgreater\\
\textless where\textgreater\\
12.5: [50,60,150,160]\\
13.5: [55,62,140,150]\\
\textless /where\textgreater\\
\textless /answer\textgreater

\label{grpo}
\end{tcolorbox}

\subsection{Additional Experiment Results}
\paragraph{Experiments on Other Datasets}
In this section, we conduct cross-dataset experiments on additional datasets to further validate the effectiveness of our model.
For the temporal reasoning evaluation, we use the AVE~\cite{tian2018audio} dataset, which consists of 10-second video clips.
For the spatial reasoning evaluation, we adopt the AVSBench-V1~\cite{zhou2022audio} dataset, which contains 5-second videos.
As shown in Tab.~\ref{tab:cross_dataset exp}, our model achieves performance comparable to VideoChat-R1~\cite{li2025videochat} in the temporal dimension and attains the best performance in the spatial dimension.
\begin{table}[h]
\centering
\small
\setlength{\tabcolsep}{1.5mm}
\begin{tabular}{l|c|c}
\toprule
\textbf{Model} & \textbf{AVE (m\_tIoU)} & \textbf{AVSBench-V1 (m\_vIoU)} \\
\midrule
Qwen2.5-VL    & 29.49  & \underline{3.79} \\
VideoChat-R1  & \textbf{40.81}  & 2.98 \\
Video-R1      & 18.37  & 1.98 \\
\rowcolor{gray!10}
\textbf{Ours}         & \underline{38.15}  & \textbf{6.84} \\
\bottomrule
\end{tabular}
\caption{Comparison of AVST-Zero on the AVE~\cite{tian2018audio} and AVSBench-V1~\cite{zhou2022audio} datasets.}
\label{tab:cross_dataset exp}
\end{table}